\documentclass{article}
\usepackage[utf8]{inputenc} % allow utf-8 input
\usepackage[T1]{fontenc}    % use 8-bit T1 fonts
\usepackage{amssymb}
\usepackage{authblk}
\usepackage{hyperref}
\usepackage{graphicx}
\usepackage{float}
\usepackage{booktabs}
\usepackage{xcolor,colortbl}
\usepackage{subcaption}
\usepackage{caption}
\usepackage{enumitem}
\usepackage{wrapfig}

\definecolor{myblue}{HTML}{1f77b4}
\definecolor{myorange}{HTML}{ff7f0e}

\usepackage[preprint]{corl_2025} % Uncomment for pre-prints (e.g., arxiv); This is like ``final'', but will remove the CORL footnote.

\title{KineDex: Learning Tactile-Informed Visuomotor Policies via Kinesthetic Teaching for Dexterous Manipulation}

\author{
  Di Zhang$^{1,5}$ \thanks{Email: 2331922@tongji.edu.cn}
  \quad
  Chengbo Yuan$^{2,5,6}$ \quad
  Chuan Wen$^3$ \quad
  Hai Zhang$^{4,5}$ \quad
  \\
  Junqiao Zhao$^{1}$ \quad
  Yang Gao$^{2,5,6}$\thanks{Corresponding at: gaoyangiiis@mail.tsinghua.edu.cn} 
  \\ \\
  $^1$Tongji University \quad 
  $^2$Tsinghua University \quad 
  $^3$Shanghai Jiao Tong University \quad
  \\
  $^4$University of Hong Kong \quad
  $^5$Shanghai Qi Zhi Institute \quad 
  $^6$Shanghai AI Lab \quad 
  \\ \\
  \href{https://dinomini00.github.io/KineDex/}{https://dinomini00.github.io/KineDex/}
}

\begin{document}
\maketitle

%===============================================================================

\begin{figure}[H]
    \centering
    \vspace{-1cm}
    \hspace*{-0.04\textwidth}
    \includegraphics[width=1.08\textwidth]{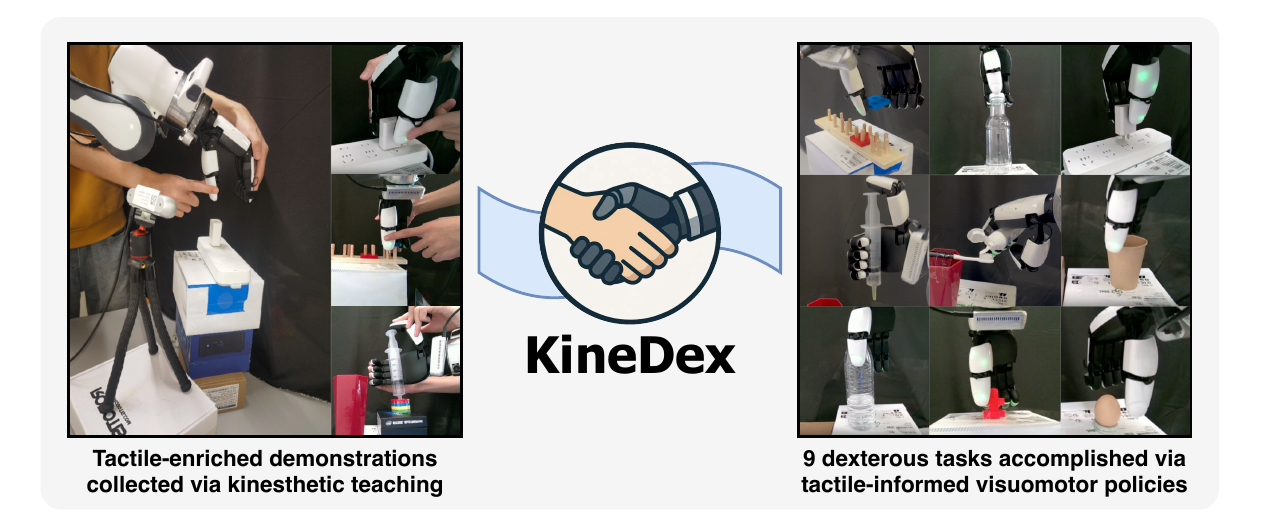}
    \caption{We present \textbf{KineDex}, a framework for collecting tactile-enriched demonstrations via kinesthetic teaching and training tactile-informed visuomotor policies for dexterous manipulation.
    }
    \label{fig:logo}
\end{figure}

\begin{abstract}
    Collecting demonstrations enriched with fine-grained tactile information is critical for dexterous manipulation, particularly in contact-rich tasks that require precise force control and physical interaction. While prior works primarily focus on teleoperation or video-based retargeting, they often suffer from kinematic mismatches and the absence of real-time tactile feedback, hindering the acquisition of high-fidelity tactile data. To mitigate this issue, we propose \textbf{KineDex}, a hand-over-hand kinesthetic teaching paradigm in which the operator’s motion is directly transferred to the dexterous hand, enabling the collection of physically grounded demonstrations enriched with accurate tactile feedback. To resolve occlusions from human hand, we apply inpainting technique to preprocess the visual observations. Based on these demonstrations, we then train a visuomotor policy using tactile-augmented inputs and implement force control during deployment for precise contact-rich manipulation. 
    We evaluate KineDex on a suite of challenging contact-rich manipulation tasks, including particularly difficult scenarios such as squeezing toothpaste onto a toothbrush, which require precise multi-finger coordination and stable force regulation.
    Across these tasks, KineDex achieves an average success rate of 74.4\%, representing a 57.7\% improvement over the variant without force control.
    Comparative experiments with teleoperation and user studies further validate the advantages of KineDex in data collection efficiency and operability. 
    Specifically, KineDex collects data over twice as fast as teleoperation across two tasks of varying difficulty, while maintaining a near-100\% success rate, compared to under 50\% for teleoperation.
\end{abstract}


\section{Introduction}
\label{sec:introduction}

Integrating tactile sensing with dexterous hands substantially improves robotic manipulation capabilities~\citep{yinTouchDexterityRotating2023,qiRotateitGeneralInHand2023,zhang2025robustdexgrasprobustdexterousgrasping,lee2025trajectoryoptimizationinhandmanipulation}, paving the way for broader deployment in daily scenarios. Despite notable advances in recent years, particularly in hardware offering increased flexibility~\citep{shawLEAPHandLowCost2023,shaw2024leap,christophORCA2025,romeroEyeSight2024} and improved tactile sensor precision~\citep{bhirangi2024anyskin,linDTactVisionBasedTactile2022,xuDTactiveVisionBasedTactile2024,wangGelSight2021}, acquiring expert-level demonstrations that incorporate high-fidelity tactile sensing remains a fundamental challenge.

Most existing demonstration collection methods focus on human hand motion retargeting. A common approach is teleoperation~\citep{gallipoliVirtual2024,wangDexCap2024,chengOpenTeleVision2024,siTilde2024,wuGELLO2024,shawVideoDex2023}, which captures the demonstrator's hand trajectories using a virtual reality headset or data gloves and maps them to robotic hands. Another line of work directly leverages egocentric videos to infer corresponding robot motions~\citep{singhHOP2024,guzeyHUDOR2024,liOKAMI2024,chenViViDex2024,li2024okamiteachinghumanoidrobots}. 
However, these methods face limitations when applied to complex manipulation tasks. First, the retargeted trajectories often fail to accurately reproduce human behavior due to the kinematic mismatch between human and robotic hands. Second, the absence of on-robot tactile feedback during teleoperation makes task performance highly dependent on the operator's expertise. To alleviate this issue, recent work has introduced exoskeleton-based systems that provide real-time haptic feedback as a proxy for touch~\citep{xuImmersive2025,zhangDOGlove2025}. However, the interaction experience still differs notably from direct physical contact, and the data collection efficiency remains comparable to that of traditional teleoperation.

Motivated by recent advances in the biomimetic design of dexterous hands~\citep{piazza2019century}, we propose \textbf{KineDex}, a framework for collecting tactile-enriched demonstrations through hand-over-hand guidance.
This paradigm offers several key advantages: 
(i) human motions can be directly applied to the robotic hand, eliminating retargeting errors; 
(ii) operators receive precise force feedback, enabling the collection of high-quality tactile data; 
(iii) data collection efficiency approaches that of direct human execution; and
(iv) the framework naturally extends to more complex hardware and challenging contact-rich dexterous manipulation tasks, addressing limitations of previous kinesthetic teaching setups~\citep{yooKineSoftLearningProprioceptive2025,chenDexForceExtractingForceinformed2025,liuForceMimicForceCentricImitation2024}.

Building upon the tactile-enriched kinesthetic demonstrations, we further leverage them for visuomotor policy training.
However, this remains challenging due to the domain shift introduced by the presence of the human hand in collected visual observations during training, but absent at inference time. 
To address this, we apply inpainting techniques~\citep{zhouProPainter2023} to remove occlusions, providing a more scalable and efficient alternative to prior trajectory replay-based methods~\citep{chenDexForceExtractingForceinformed2025}.
To better perform contact-rich manipulation, we incorporate tactile sensing to enrich the policy inputs. 
Moreover, the policy is trained to predict target fingertip forces alongside target joint positions, which we refer to as force-informed actions, enabling KineDex to achieve accurate force control during execution.

We conduct extensive experiments on nine contact-rich manipulation tasks, as illustrated in Figure~\ref{fig:logo}. KineDex demonstrates robust and precise control, achieving an average success rate of 74.4\% across all tasks and outperforming the variant without force control by 57.7\%. 
In addition, we find that tactile sensing is particularly important for contact-intensive tasks such as \textit{Cap Twisting}, \textit{Toothpaste Squeezing}, and \textit{Syringe Pressing}. Experimental results show that KineDex outperforms the variant without tactile sensing by 26.7\% on these tasks. 
We conduct comparative experiments with teleoperation to further assess the efficiency of KineDex. The results show that KineDex achieves more than twice the data collection speed across two tasks of varying difficulty, while maintaining a success rate close to 100\%.
In contrast, teleoperation yields an average success rate below 50\% and requires significantly more time to collect the same amount of data. User study feedback additionally confirms the advantages of KineDex over teleoperation, with participants reporting a more intuitive and efficient experience when using kinesthetic teaching.

In summary, we present \textbf{KineDex}, a framework for dexterous manipulation that integrates kinesthetic data collection with tactile-informed visuomotor policy training. Through empirical evaluations and user studies, we demonstrate that kinesthetic teaching provides a more effective and efficient alternative to teleoperation for collecting high-quality demonstrations with dexterous hands. Furthermore, by leveraging tactile-augmented demonstrations, we train a tactile-informed policy that incorporates force control during inference, enabling successful execution across nine distinct contact-rich manipulation tasks.

\section{Related Work}
\label{sec:related_work}

\subsection{Collecting Demonstrations with Dexterous Hands}

Most existing methods for collecting demonstrations with dexterous hands rely on retargeting human hand motion via teleoperation~\citep{qinAnyTeleop2024,handaDexPilot2019,iyerOPEN2024,arunachalamHoloDexTeachingDexterity2022,ding2024bunnyvisionprorealtimebimanualdexterous} or using video data~\citep{singhHOP2024,guzeyHUDOR2024,liOKAMI2024,chenViViDex2024,li2024okamiteachinghumanoidrobots}.
However, both approaches share a key limitation: the operator lacks real-time tactile feedback, adversely impacting the efficiency and success rate of data collection.
To mitigate this, some studies have introduced exoskeleton systems~\citep{chaoExoViHaCrossPlatformExoskeleton2025,xuImmersive2025,zhangDOGlove2025} that simulate haptic feedback during interaction.
Nevertheless, such approximations differ fundamentally from the actual tactile sensations experienced through direct physical contact.

Kinesthetic teaching~\citep{houAdaptiveCompliancePolicy2024,liuForceMimicForceCentricImitation2024,ablettMultimodalForceMatchedImitation2025,weiWearable2023,chenDexForceExtractingForceinformed2025} enables operators to physically manipulate the robotic hardware and receive real-time force feedback.
HIRO~\citep{weiWearable2023} introduces a hand-over-hand teaching paradigm, where the demonstrator grasps the robotic hand to experience accurate force feedback, but does not record tactile data, limiting its applicability to complex tasks.
The most closely related work to ours is DexForce~\citep{chenDexForceExtractingForceinformed2025}, which collects tactile-augmented demonstrations via kinesthetic teaching but relies on simpler hardware with fewer degrees of freedom and evaluates only on basic tasks such as grasping or flipping.
Moreover, due to visual occlusions from the demonstrator’s hand, DexForce requires replaying kinesthetic trajectories to obtain clean observations—a process that becomes infeasible for longer-horizon tasks.

\subsection{Learning Dexterous Manipulation from Human Demonstrations}

Learning from expert human demonstrations~\citep{mandlekarLearning2021,chang2023oneshotvisualimitationattributed,yu2019one,johns2021coarsetofineimitationlearningrobot,chiDiffusionPolicyVisuomotor2024,ze3DDiffusionPolicy2024} enables embodied agents to acquire autonomous manipulation skills. Diffusion Policy~\citep{chiDiffusionPolicyVisuomotor2024}, which applies diffusion models to imitation learning, has shown strong capability in capturing the multimodal structure of demonstration data. Extensions such as 3D Diffusion Policy~\citep{ze3DDiffusionPolicy2024} integrate point cloud information into the perception pipeline, allowing for richer environmental representations. Other works have explored the integration of tactile sensing~\citep{sunVTAOBiManipMaskedVisualTactileAction2025,guzeyDexterity2023,guzeySee2023}, enabling policies to perform more fine-grained manipulation. In our work, we adopt Diffusion Policy as the backbone for policy learning, augmenting its observation space with tactile sensing and applying force control during inference to ensure precision and stability in contact-rich manipulation tasks.
\section{Method}

\subsection{Problem Formulation}
\label{sec:formulation}

In this work, we aim to train tactile-informed visuomotor policies from kinesthetic demonstrations via imitation learning, and to implement force control at inference time for executing contact-rich manipulation tasks. The observation space and learning targets are defined as follows.

\textbf{Observation Space.}
At each time step $t$, the policy receives multi-modal observations comprising:
(i) RGB images captured from $N_o$ multi-view cameras, denoted as $o_t \in \mathbb{R}^{N_o \times H \times W \times 3}$;
(ii) tactile vectors obtained from $N_q$ sensing points on each of the five fingertips, represented as $q_t \in \mathbb{R}^{5 \times N_q}$;
(iii) proprioceptive signals from $N_x$ joints of the robotic hardware, given by $x_t \in \mathbb{R}^{N_x}$.

\textbf{Action Space.} 
To enable precise force control during inference, the policy predicts not only the target joint positions $x_d \in \mathbb{R}^{N_x}$, but also the $N_f$-dimensional target fingertip forces $f_d \in \mathbb{R}^{5 \times N_f}$, resulting in a force-informed action denoted by $a_t = \{ x_d, f_d \}$. While $x_d$ specifies the baseline configuration of the hand, $f_d$ modulates the contact forces via a modified PD controller (introduced in Section~\ref{sec:force_control}), allowing the fingers to actively track the desired contact forces during execution.

\begin{figure}[!t]
    \centering
    \hspace*{-0.02\textwidth}
    \includegraphics[width=1.03\textwidth]{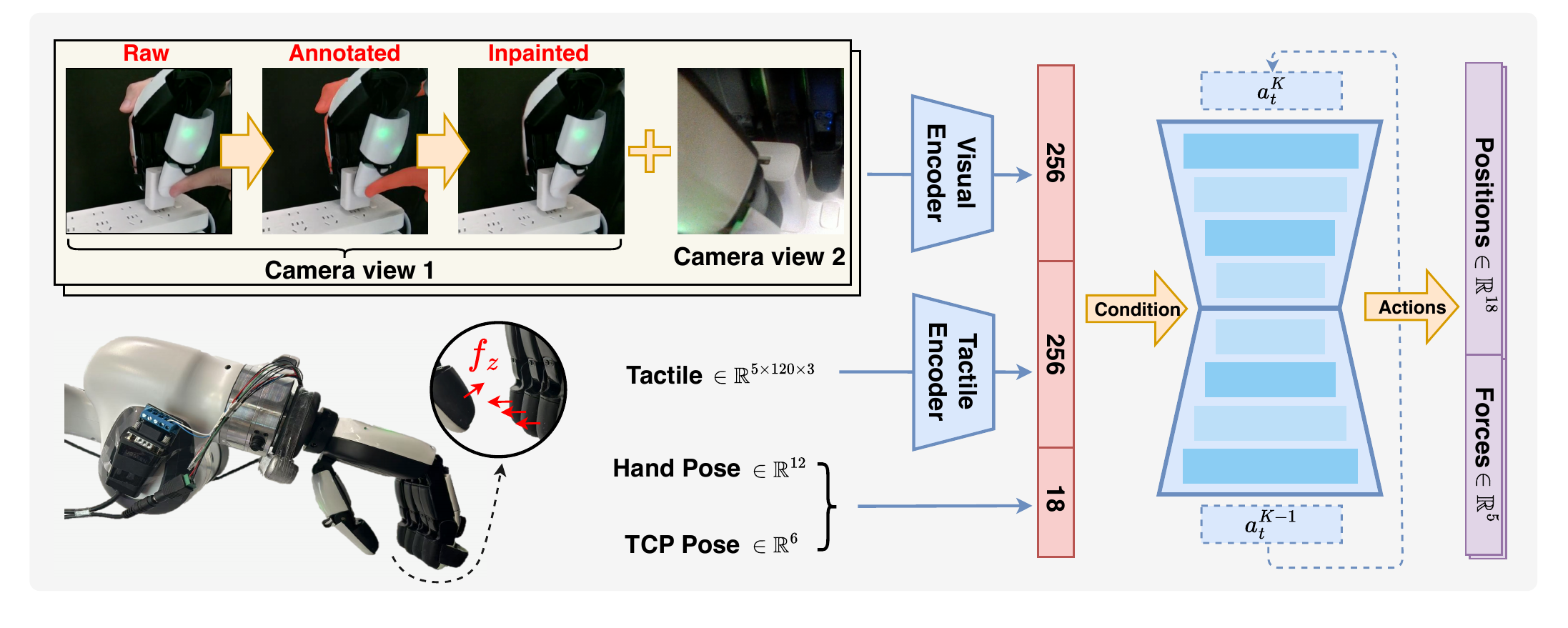}
    \vspace{-0.6cm}
    \caption{Overview of the \textbf{KineDex} framework. KineDex collects tactile-enriched demonstrations via kinesthetic teaching, where visual occlusions from the operator’s hand are removed through inpainting before policy training. The learned policy takes visual and tactile inputs to predict joint positions and contact forces, which are executed with force control for robust manipulation.}
    \vspace{-0.5cm}
    \label{fig:pipline}
\end{figure}

\subsection{Kinesthetic Data Collection}
\label{sec:hardware}

The general hardware setup of \textbf{KineDex} consists of a robotic arm equipped with a dexterous hand, as illustrated in Figure~\ref{fig:logo}. In addition, we employ two RGB cameras to capture visual observations: one is mounted in front of the workspace to provide a global view of the scene, and the other is wrist-mounted on the end-effector to enable close-range perception of the manipulation area.

The core idea of KineDex data collection system is to allow operators to “wear” the dexterous hand while moving freely to perform precise contact-rich manipulation in real time.
To enable this hand-over-hand control, we attach ring-shaped straps to the dorsal sides of the four non-thumb fingers of the robotic hand, allowing the operator to guide the hand as if wearing a glove. 
This physical coupling ensures that contact forces experienced during motion are immediately transmitted to the operator’s hand, providing natural haptic feedback throughout the demonstration. 
Due to morphological differences between human and robotic hands, the operator controls the thumb separately with their left hand, while guiding the remaining fingers with their right hand as described above. Examples of kinesthetic demonstrations are shown in Figure~\ref{fig:compare_with_teleop}.

During kinesthetic teaching, the following data modalities are recorded for each demonstration:
\begin{itemize}[leftmargin=0.5cm]
    \item \textit{Visual observations}: RGB images captured from the front-facing and wrist-mounted cameras. As raw observations may contain the operator’s body, we apply an inpainting strategy to address this issue, as detailed in Section~\ref{sec:policy_learning}.
    \item \textit{Proprioception}: The robot arm’s end-effector pose and the dexterous hand’s joint positions.
    \item \textit{Tactile sensing}: Per-finger tactile measurements, with each finger equipped with multiple sensing points that record localized contact forces, forming a dense tactile sensing matrix.
    \item \textit{Fingertip force}: A 3D force vector $\mathbf{f} = (f_x, f_y, f_z)$ for each fingertip, where each component represents the force along the corresponding axis, computed by aggregating the localized forces from all tactile sensing points.
\end{itemize}

\subsection{Policy Learning}
\label{sec:policy_learning}

The raw kinesthetic demonstrations collected by the system cannot be directly used for visuomotor policy learning, as the front-facing camera inevitably captures the operator’s body during interaction. Training on such data introduces a significant out-of-distribution(OOD) shift at inference time, when the human body is no longer present in the scene (as shown in Table~\ref{tab:results}). Motivated by recent advances in human-to-robot data editing~\citep{chenRoViAug2024,lepert2025phantomtrainingrobotsrobots}, we adopt an inpainting-based approach to remove the operator’s body from the visual observations.

For raw kinesthetic demonstrations, we first apply Grounded-SAM~\citep{renGrounded2024} to extract masks of the operator’s body parts from the video frames. These frames, along with their corresponding masks, are then passed to the ProPainter~\citep{zhouProPainter2023} model to inpaint the occluded human body regions. An example of the data preprocessing pipeline is illustrated in Figure~\ref{fig:pipline}. Although the inpainting model is not pretrained on robot-specific data and may not achieve perfect removal, our experiments demonstrate that the resulting demonstrations are sufficient for training high-performance policies.

Using the preprocessed demonstrations, we train Diffusion Policy~\citep{chiDiffusionPolicyVisuomotor2024} conditioned on inpainted visual observations, tactile sensing, and proprioception to predict force-informed actions, modeled as $p(x_d, f_d \mid o_t, q_t, x_t)$. Specifically, for the calculated fingertip force vector $\mathbf{f} = (f_x, f_y, f_z)$, we supervise policy training using the normal force component $f_z$, which corresponds to the primary axis along which the fingertip can actively exert force, as illustrated in Figure~\ref{fig:pipline}. 

At inference time, the policy produces action chunks~\citep{zhaoLearning2023} to enable smoother control. Each chunk specifies the desired joint positions and fingertip contact forces, which are executed using the force control strategy described below. Further details of the network architecture and policy training configurations are provided in Appendix~\ref{app:policy}.

\subsection{Force Control}
\label{sec:force_control}

In conventional setups, robots are typically operated under position control~\citep{siciliano2008springer}, where the control signal $u$ is computed by a PD controller~\citep{johnson2005pid} based on the joint position and velocity errors relative to the target joint positions $x_d$:
\begin{equation}
    u = K_p(x_d - x) + K_d(\dot{x_d} - \dot{x}),
\end{equation}
where $K_p$ and $K_d$ denote the proportional and derivative gains, respectively.

However, relying solely on position control may be insufficient for certain precise, contact-rich tasks, such as \textit{cap twisting} or \textit{toothpaste squeezing}.
This is due to policies that only track target joint positions results in merely contacting the object’s surface without applying meaningful forces, as the recorded fingertip positions remain unchanged regardless of the forces applied during kinesthetic teaching. 
This discrepancy often leads to unstable grasps, slipping, or ineffective manipulation.

To address this limitation, we exploit the following physical property: when a fingertip contacts an object, any nonzero position error continuously generates pressure against the surface due to the object’s resistance, with larger errors producing greater forces, as if the target position lies inside the object. 
We refer to this virtual displacement as \textbf{the force-informed target position}.
To compute it, we utilize the predicted fingertip forces $f_d$ from the trained policy, which are directed orthogonal to the contact surface and align with the primary motion axis of the finger joints.
Specifically, for each finger, let $x^{tip}$ and $x^{base}$ denote the current positions of the fingertip and base joints, respectively. The force-informed target positions, $x_d^{tip}$ and $x_d^{base}$, are then computed as:
\begin{equation}
    \left\{
    \renewcommand{\arraystretch}{1.5}
    \begin{array}{l}
    x_{d}^{tip} = x^{tip} + K^{tip} \cdot f_{d} \\
    x_{d}^{base} = x^{base} + K^{base} \cdot f_{d}
    \end{array}
    \right.
\end{equation}
Here, $K^{tip}$ and $K^{base}$ are two hyperparameters that determine the motion stiffness at the fingertip and base joints. They are tuned to ensure that the execution faithfully tracks the predicted forces and are kept fixed across different tasks. With this force control strategy, KineDex can precisely track the target fingertip forces predicted by the trained policy, thereby achieving stable and force-informed control during execution.
\section{Experiments}
\label{sec:experiments}

In this section, we investigate the effectiveness of kinesthetic demonstrations for training visuomotor policies across a range of contact-rich dexterous manipulation tasks.
We further evaluate the efficiency and practicality of kinesthetic teaching through comparative experiments and a user study, highlighting its advantages over teleoperation-based approaches.

To this end, we design a suite of nine tasks that emphasize precise force control, multi-finger coordination, and interaction with everyday objects.
These tasks span a range of dexterous skills, including challenging scenarios such as squeezing toothpaste onto a toothbrush, which requires continuous and fine-grained pressure modulation; and pressing a syringe, which demands stable unimanual actuation and a coordinated grip to prevent slippage or misalignment.
Detailed task descriptions are provided in Appendix~\ref{app:task}.

\begin{table}[!t]
    \centering
    \vspace{-0.3cm}
    \caption{Number of successful trials (out of 20) \textbf{during inference} for different methods.}
    \resizebox{\textwidth}{!}{%
        \begin{tabular}{lccccc}
            \toprule
            \textbf{Method} & \textit{Bottle Picking} & \textit{Cup Picking} & \textit{Egg Picking} & \textit{Cap Twisting} & \textit{Nut Tightening}  \\
            \midrule
            \midrule
            \textbf{KineDex} & 17 & 20 & 17 & 15 & 16  \\
            w/o Force Control & 0 & 16 & 5 & 2 & 7  \\
            w/o Tactile Input & 15 & 17 & 18 & 10 & 12  \\
            w/o Inpainting & 0 & 0 & 0 & 0 & 0  \\
            \bottomrule
        \end{tabular}%
    }
    \\[0.1cm]
    \resizebox{\textwidth}{!}{%
        \begin{tabular}{lcccc}
            \toprule
            \textbf{Method} & \textit{Peg Insertion} & \textit{Charger Plugging} & \textit{Toothpaste Squeezing} & \textit{Syringe Pressing} \\
            \midrule
            \midrule
            \textbf{KineDex} & 15 & 12 & 9 & 13  \\
            w/o Force Control & 0 & 0 & 0 & 0  \\
            w/o Tactile Input & 16 & 10 & 3 & 8  \\
            w/o Inpainting & 0 & 0 & 0 & 0 \\
            \bottomrule
        \end{tabular}%
    }
    %\vspace{-0.2cm}
    \label{tab:results}
\end{table}

\begin{figure}[!t]
    \centering
    \begin{subfigure}{0.495\textwidth}
        \centering
        \includegraphics[width=\linewidth]{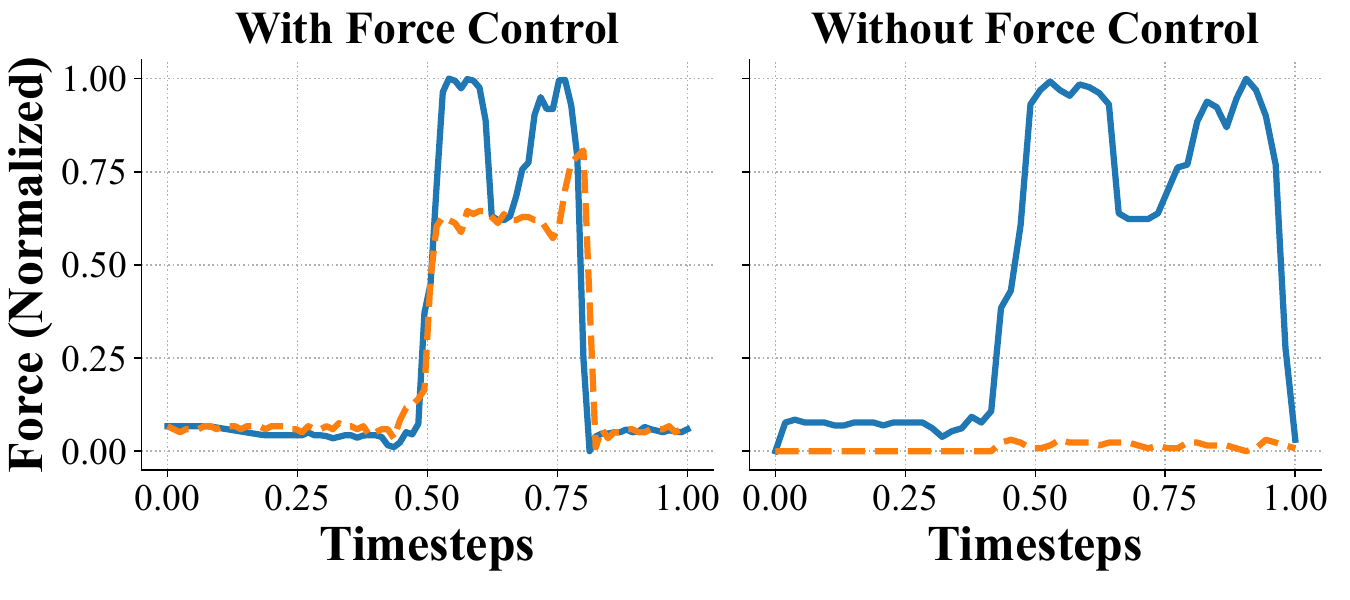}
        \vspace{-0.5cm}
        \caption{\textit{Charger Plugging}.}
    \end{subfigure}
    \hfill
    \begin{subfigure}{0.495\textwidth}
        \centering
        \includegraphics[width=\linewidth]{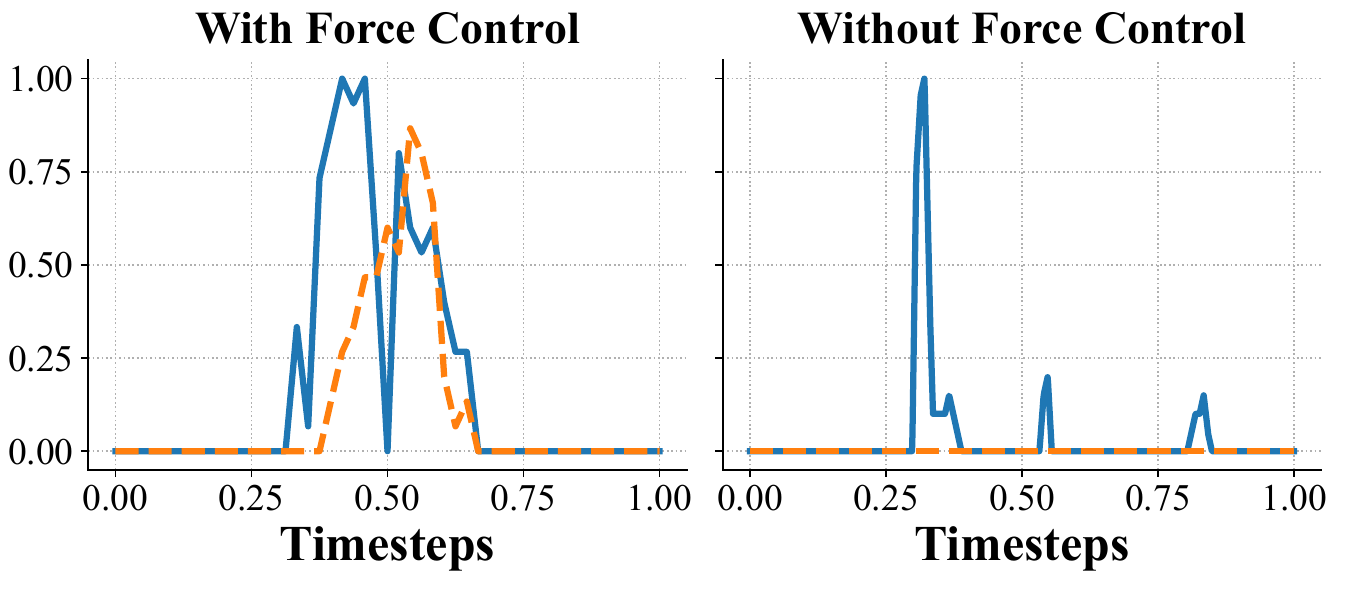}
        \vspace{-0.5cm}
        \caption{\textit{Egg Picking}.}
    \end{subfigure}
    \caption{Visualization of \textcolor{myblue}{predicted} and \textcolor{myorange}{sensed} forces at the thumb during task execution, comparing the force-informed policy and the variant without force control.}
    \vspace{-0.5cm}
    \label{fig:visualization}
\end{figure}

\subsection{Performance Evaluation}

\textbf{Hardware Setup.}
For this set of experiments, we implement KineDex using a Franka Emika Panda robotic arm equipped with a Robotera XHand1\footnote{\url{https://www.robotera.com/goods/2.html}} dexterous hand.
Each finger on the XHand1 has two joints, while the thumb and index finger include an additional rotational joint, resulting in a total of 12 degrees of freedom.
Each finger is equipped with 120 tactile sensing points.

\textbf{Baselines.} 
We compare KineDex against three ablated variants:
(i) \textit{w/o Force Control}, which disables force control during inference while keeping the training setup unchanged.
(ii) \textit{w/o Tactile Input}, which removes tactile sensing from the policy inputs during training; however, the policy still predicts target fingertip forces, which are executed using the same force control strategy.
(iii) \textit{w/o Inpainting}, which omits the inpainting preprocessing step for kinesthetic demonstrations.

We evaluate performance by conducting 20 trials per task, with results summarized in Table~\ref{tab:results}. KineDex achieves over 70\% success rates on most tasks, and nearly 100\% on common pick-and-place scenarios such as \textit{Bottle Picking} and \textit{Cup Picking}.
While performance is slightly lower on the final three, more challenging tasks, the average success rates still exceed 50\%.
This drop is likely due to the increased demands for fine-grained localization and contact reasoning, which may exceed the representational capacity of the current policy inputs.
Despite these challenges, the results demonstrate that kinesthetic demonstrations effectively support visuomotor policy learning across a wide range of daily manipulation tasks, owing to their natural alignment with human behavior and the availability of accurate tactile and force feedback.

\begin{table}[!t]
    \centering
    %\vspace{-0.5cm}
    \caption{Number of successful trials (out of 20) \textbf{during data collection} for different methods.}
    %\vspace{0.25cm}
    \resizebox{\textwidth}{!}{%
        \begin{tabular}{lccccc}
            \toprule
            \textbf{Method} & \textit{Bottle Picking} & \textit{Cup Picking} & \textit{Cap Twisting} & \textit{Charger Plugging} & \textit{Syringe Pressing}  \\
            \midrule
            \midrule
            \textbf{KineDex} & 20 & 20 & 20 & 18 & 20  \\
            Teleoperation & 19 & 9 & 7 & 0 & 4  \\
            \bottomrule
        \end{tabular}%
    }
    \vspace{-0.35cm}
    \label{tab:results2}
\end{table}

\begin{figure}[!t]
    \centering
    \hspace*{-0.035\textwidth}
    \includegraphics[width=1.06\textwidth]{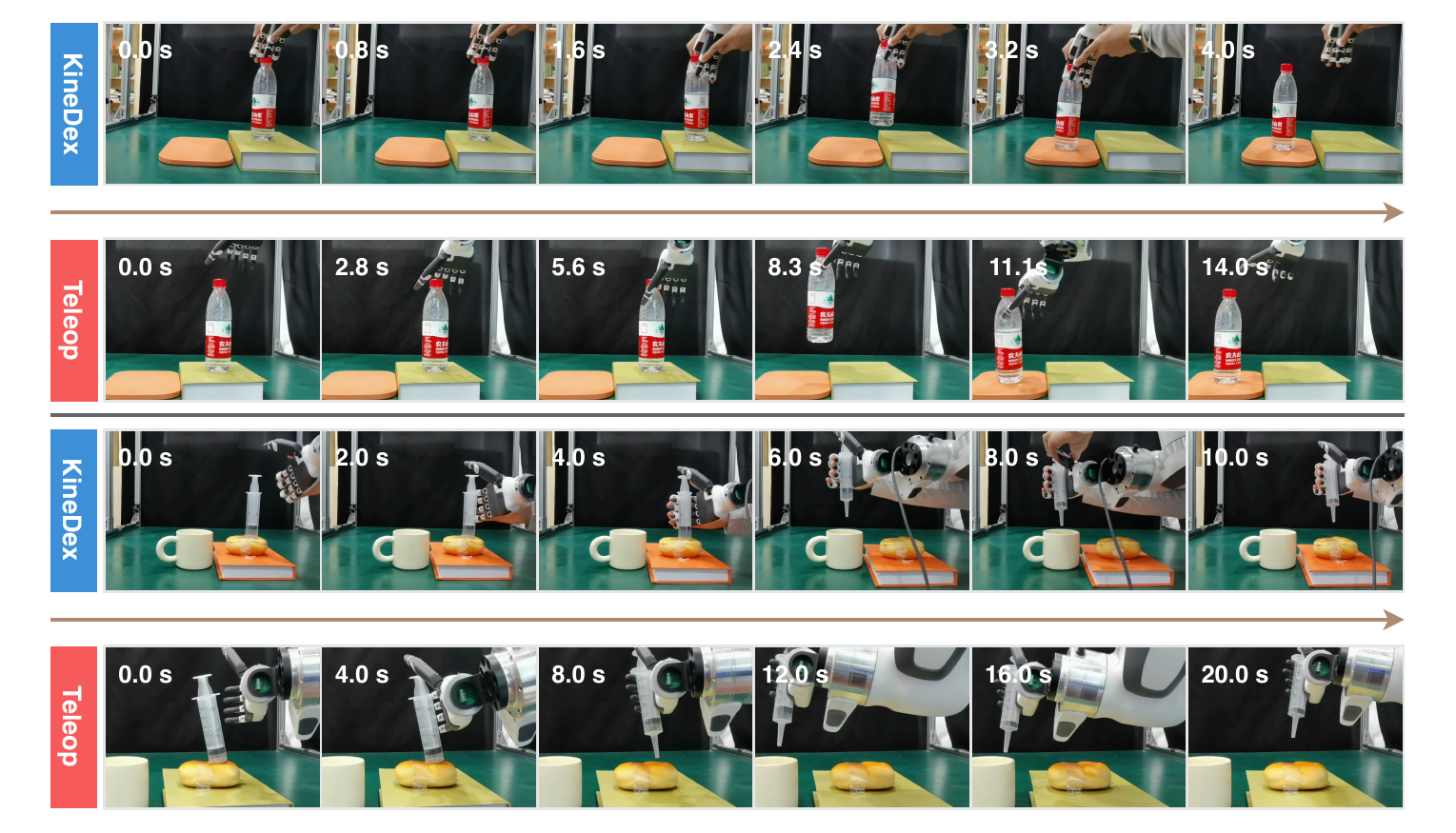}
    \vspace{-0.5cm}
    \caption{Comparison of demonstration collection time between \textbf{KineDex} and teleoperation on the \textit{Bottle Picking} and \textit{Syringe Pressing}.}
    \vspace{-0.6cm}
    \label{fig:compare_with_teleop}
\end{figure}

\textbf{The ablation results without force control} underscore the importance of incorporating both force measurements during training and force control during inference.
In this setting, the average success rate across all tasks drops to just 16.7\%, with even simple tasks such as \textit{Bottle Picking} rarely completed successfully.
Without force control, the robotic hand often merely contacts the object surface without applying sufficient pressure, leading to frequent failures in contact-rich tasks.
To further examine this effect, Figure~\ref{fig:visualization} visualizes the fingertip-object contact forces during inference.
KineDex accurately tracks the desired contact forces predicted by the policy, exhibiting similar magnitudes and temporal patterns.
In contrast, without force control, the executed force remains flat and fails to follow the predicted targets, indicating a breakdown in physical interaction quality.

\textbf{When the policy is trained without tactile input,} performance exhibits a moderate drop on most relatively simple pick-and-place tasks.
This outcome is expected, as visual information alone is often sufficient to estimate required forces in many scenarios.
However, for more contact-intensive tasks such as \textit{Cap Twisting}, \textit{Toothpaste Squeezing}, and \textit{Syringe Pressing}, removing tactile input leads to a significant performance decline, with average success rates decreasing by 26.7\%.
This suggests that in scenarios with severe visual occlusion or tasks heavily reliant on contact feedback, tactile sensing serves as an effective auxiliary modality that substantially improves task success.

\textbf{The variant without inpainting} yields a zero success rate across all tasks and exhibits unreasonable behaviors during execution.
These results confirm that raw kinesthetic demonstrations are infeasible for direct policy training due to out-of-distribution visual observations, and that inpainting provides an effective strategy for mitigating this issue.

\subsection{Efficiency Evaluation}
\label{sec:efficiency}

We further validate the advantages of KineDex over teleoperation for data collection through comparative experiments.
We replicate the Open-TeleVision~\citep{chengOpenTeleVision2024} setup to construct a teleoperation system using a Franka Emika Panda arm equipped with an Inspire Hand\footnote{\url{https://www.inspire-robots.com/product/frwz/}}, and track the operator’s hand motion using the Meta Quest 3 headset\footnote{\url{https://www.meta.com/quest/quest-3/}}.
To ensure a fair comparison, we implement KineDex using the same Inspire Hand in this set of experiments.
Detailed setup is provided in Appendix~\ref{app:teleop}.
% This setup also enables us to assess the applicability of KineDex across different dexterous hand hardware platforms.

We evaluate five tasks that are feasible under teleoperation and report the demonstration success rates during data collection.
As shown in Table~\ref{tab:results2}, teleoperation achieves an average success rate of 39\%, whereas KineDex consistently achieves near-perfect success across all tasks.
The absence of real-time tactile feedback significantly impairs teleoperation performance, particularly in tasks such as \textit{Cup Picking}, where the operator frequently crushes the paper cup due to excessive force.
Moreover, because the operator views the scene through a remote camera in the VR interface, the resulting unnatural visual feedback introduces additional challenges, especially for fine-grained manipulation such as \textit{Syringe Pressing}.
These results suggest that teleoperation requires greater operator expertise and repeated trial-and-error to produce high-quality demonstrations, leading to significantly lower data collection efficiency compared to  KineDex.

In addition to success rates, we also measure the time required to collect demonstrations with both methods.
As shown in Figure~\ref{fig:compare_with_teleop}, the improvement in efficiency is substantial: on the complex task of \textit{Syringe Pressing}, KineDex completes each demonstration in roughly half the time required by teleoperation; on the simpler task of \textit{Bottle Picking}, it takes less than one-third of the time.
This efficiency gap is primarily due to the additional time required for the operator to adjust to precise hand poses, as well as the inherent latency and limited responsiveness of the teleoperation system.

\begin{figure}[!t]
    \centering
    % \vspace{-0.3cm}
    
    \begin{subfigure}[t]{0.24\textwidth}
        \centering
        \includegraphics[width=\linewidth]{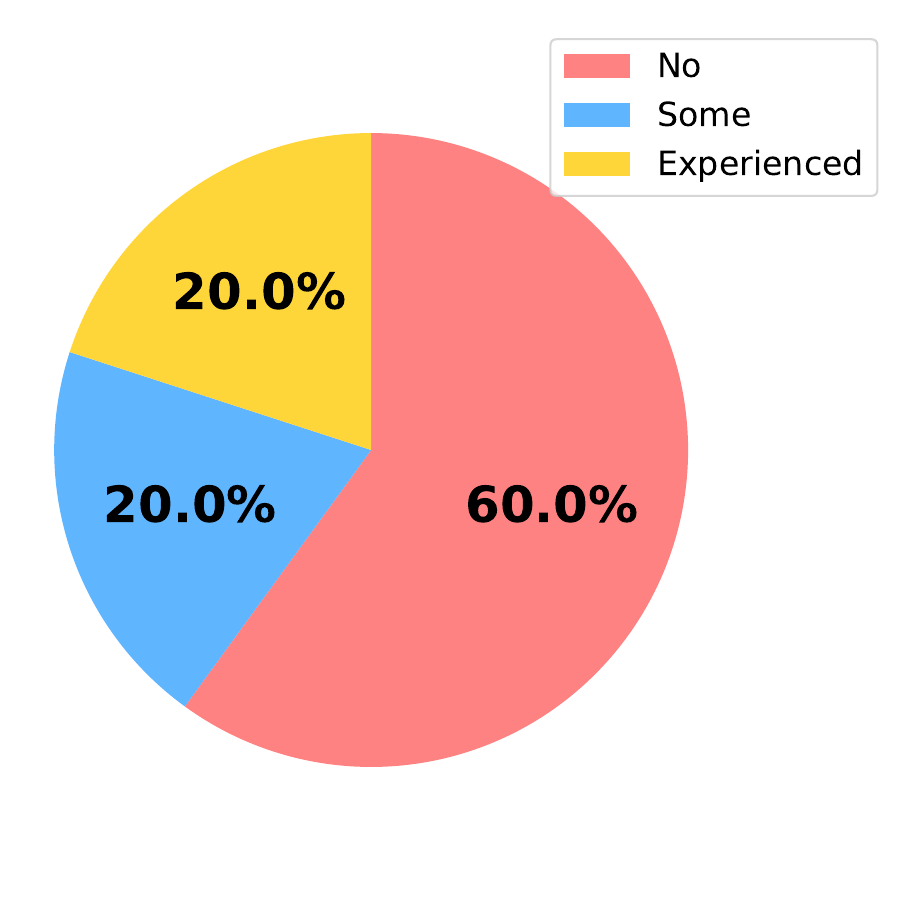}
        \vspace{-0.8cm}
        \caption{Do you have experience with teleoperation?}
    \end{subfigure}
    \hspace{0.000\textwidth}
    \begin{subfigure}[t]{0.24\textwidth}
        \centering
        \includegraphics[width=\linewidth]{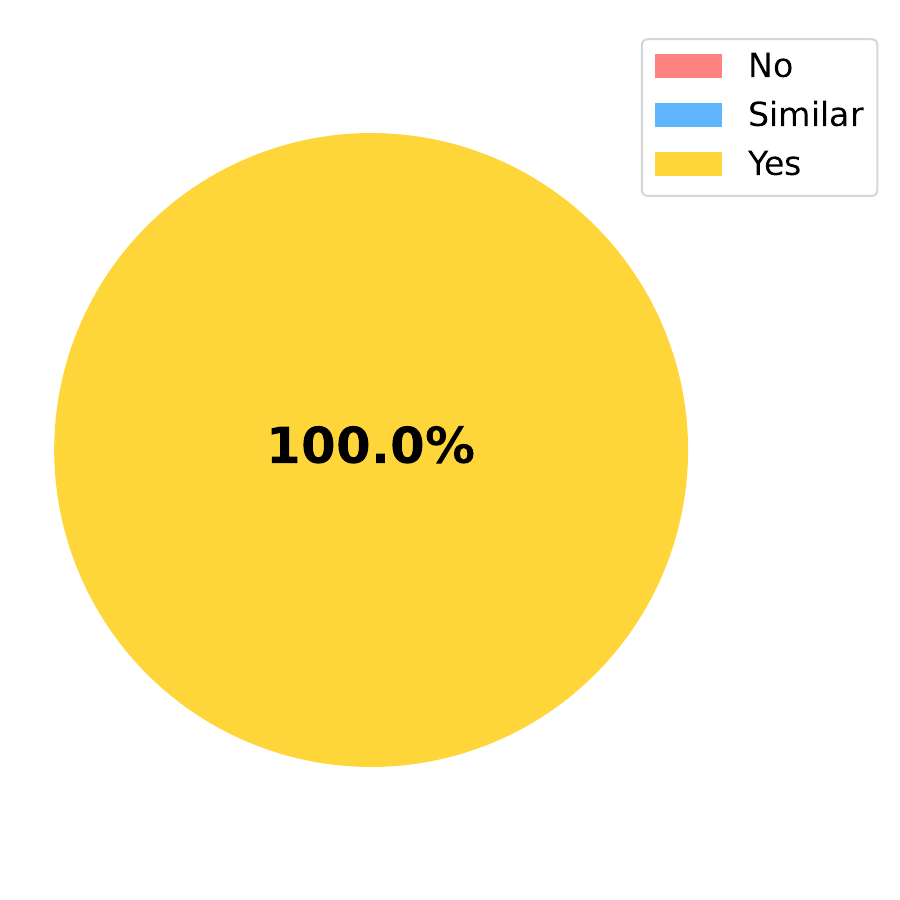}
        \vspace{-0.8cm}
        \caption{Does \textbf{KineDex} help collect more accurate tactile data?}
    \end{subfigure}
    \hspace{0.005\textwidth}
    \begin{subfigure}[t]{0.24\textwidth}
        \centering
        \includegraphics[width=\linewidth]{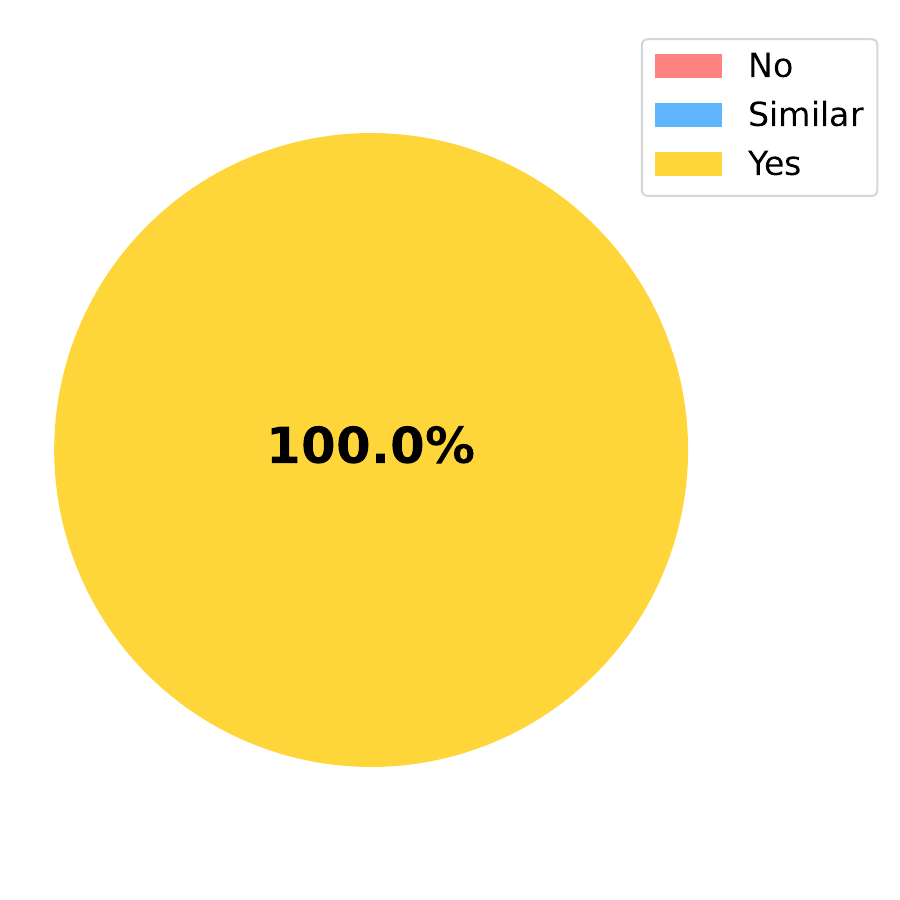}
        \vspace{-0.8cm}
        \caption{Does \textbf{KineDex} help collect demonstrations for more complex tasks?}
    \end{subfigure}
    \hspace{0.000\textwidth}
    \begin{subfigure}[t]{0.24\textwidth}
        \centering
        \includegraphics[width=\linewidth]{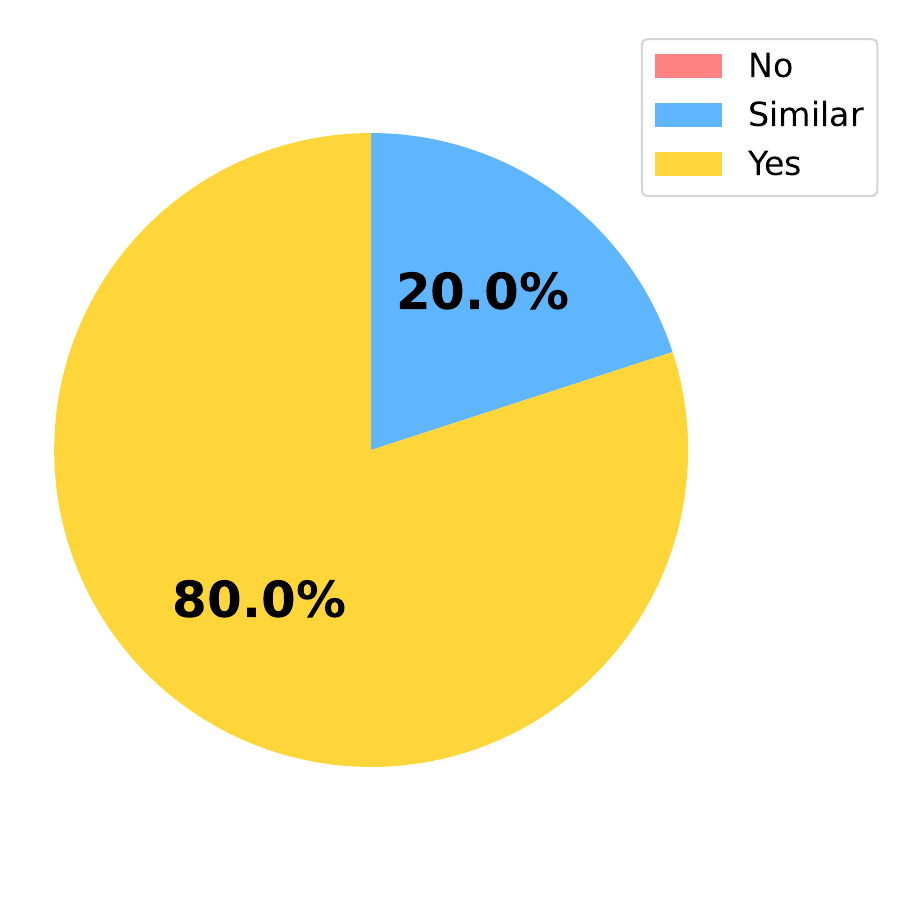}
        \vspace{-0.8cm}
        \caption{Is \textbf{KineDex} easier to use?}
    \end{subfigure}

    \caption{Summary of user study results. Five participants used both the teleoperation system and \textbf{KineDex} to collect demonstrations. Pie charts summarize their feedback on key evaluation criteria.}
    \label{fig:user_study}
    \vspace{-0.5cm}
\end{figure}

\subsection{User Study}

To provide a more comprehensive evaluation of KineDex, we conduct a user study using the setup described in Appendix~\ref{app:user}. Participants used both KineDex and the teleoperation system, and subsequently provided feedback on their perceived effectiveness and ease of use. The results, summarized in Figure~\ref{fig:user_study}, indicate that KineDex outperforms teleoperation across all evaluation criteria.
Specifically, all participants agreed that KineDex enables more accurate tactile data collection and is better suited for complex manipulation tasks, while most found it easier to use than teleoperation.

\section{Conclusion}
\label{sec:conclusion}

We present \textbf{KineDex}, a framework for collecting tactile-enriched demonstrations and training visuomotor policies for dexterous manipulation.
Through experiments on nine contact-rich manipulation tasks, we demonstrate the feasibility of kinesthetic teaching for data collection, and highlight the critical roles of tactile sensing and force control in addressing complex manipulation challenges.
Comparative studies with teleoperation further reveal that our approach offers substantial advantages in both data collection efficiency and user experience.
We hope that KineDex provides a new perspective for future research on scalable and effective data collection for dexterous robotic hands.
\section{Limitations}
\label{sec:limitations}

Based on experimental results and user study feedback, we summarize the limitations of \textbf{KineDex} as follows.
We view these limitations as promising directions for future work and hope they inspire further advancements in subsequent research:
\begin{itemize}[leftmargin=0.5cm]
    \item Although our results show that inpainting occluded regions is sufficient for training visuomotor policies, its effectiveness may degrade under more severe occlusions. This issue could potentially be mitigated by fine-tuning the inpainting model on robot-specific data.
    
    \item The current setup requires two human hands to control a single dexterous hand due to the morphological mismatch between the robotic and human thumbs, limiting scalability for bimanual demonstrations. Future work could explore more biomimetic hardware designs to enable single-handed kinesthetic teaching.
\end{itemize}
\bibliography{main}

\clearpage
\appendix

\section{Task Design}
\label{app:task}

\begin{figure}
    \centering
    
    \begin{subfigure}[t]{1\textwidth}
        \centering
        \includegraphics[width=\linewidth]{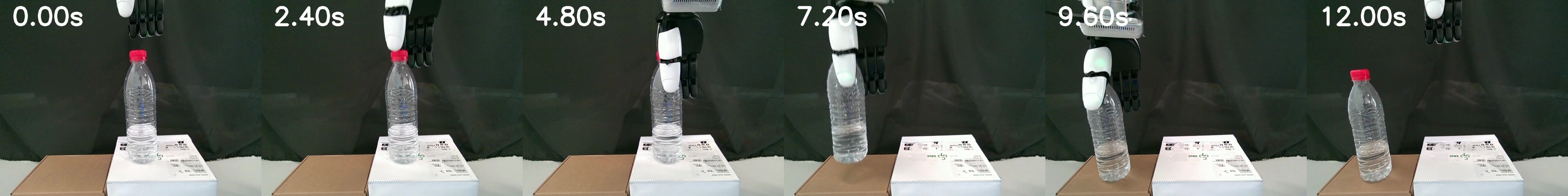}
        \caption{Bottle Picking}
    \end{subfigure}

    \begin{subfigure}[t]{1\textwidth}
        \centering
        \includegraphics[width=\linewidth]{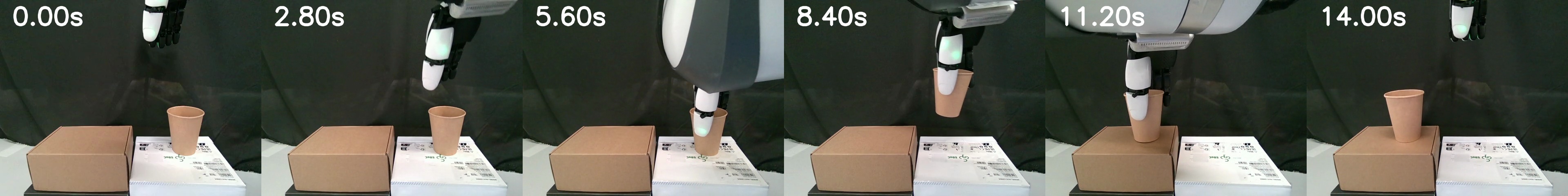}
        \caption{Cup Picking}
    \end{subfigure}

    \begin{subfigure}[t]{1\textwidth}
        \centering
        \includegraphics[width=\linewidth]{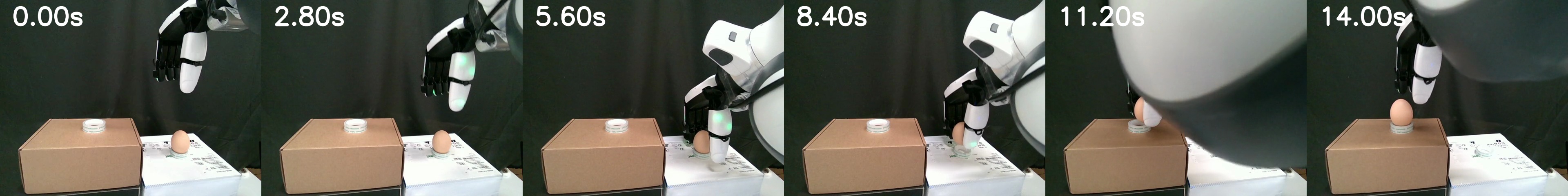}
        \caption{Egg Picking}
    \end{subfigure}

    \begin{subfigure}[t]{1\textwidth}
        \centering
        \includegraphics[width=\linewidth]{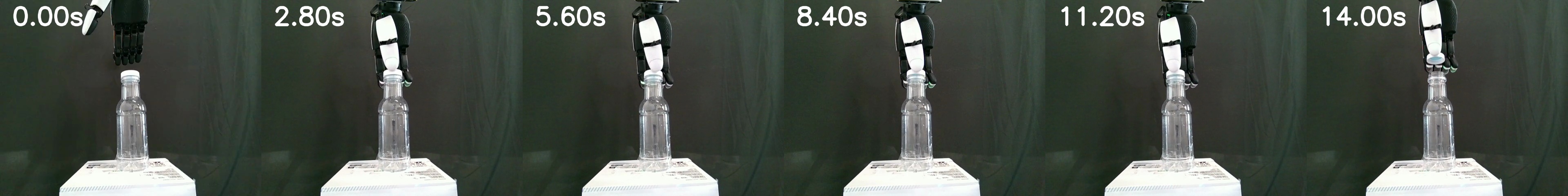}
        \caption{Cap Twisting}
    \end{subfigure}

    \begin{subfigure}[t]{1\textwidth}
        \centering
        \includegraphics[width=\linewidth]{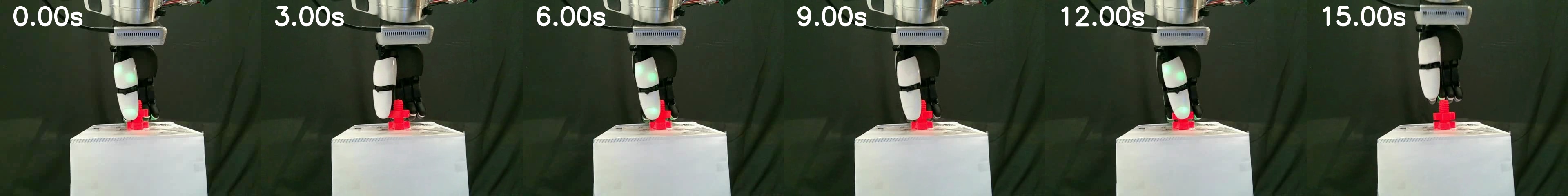}
        \caption{Nut Tightening}
    \end{subfigure}

    \begin{subfigure}[t]{1\textwidth}
        \centering
        \includegraphics[width=\linewidth]{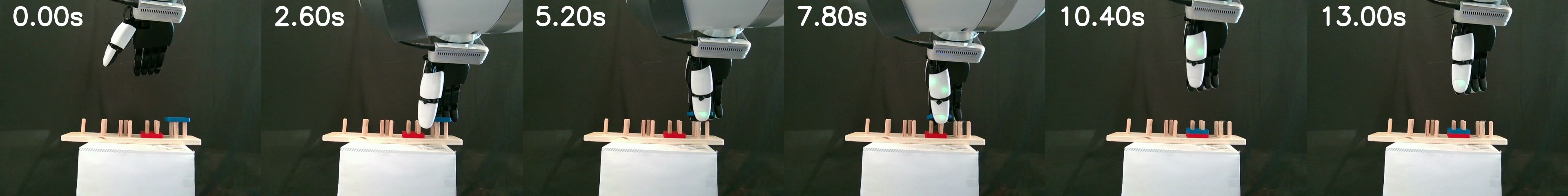}
        \caption{Peg Insertion}
    \end{subfigure}

    \begin{subfigure}[t]{1\textwidth}
        \centering
        \includegraphics[width=\linewidth]{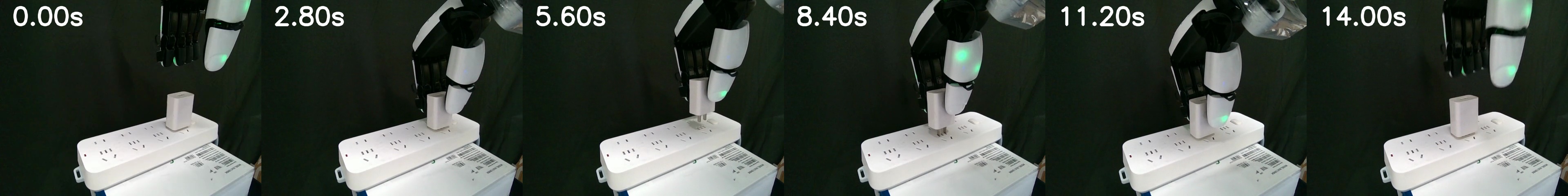}
        \caption{Charger Plugging}
    \end{subfigure}

    \begin{subfigure}[t]{1\textwidth}
        \centering
        \includegraphics[width=\linewidth]{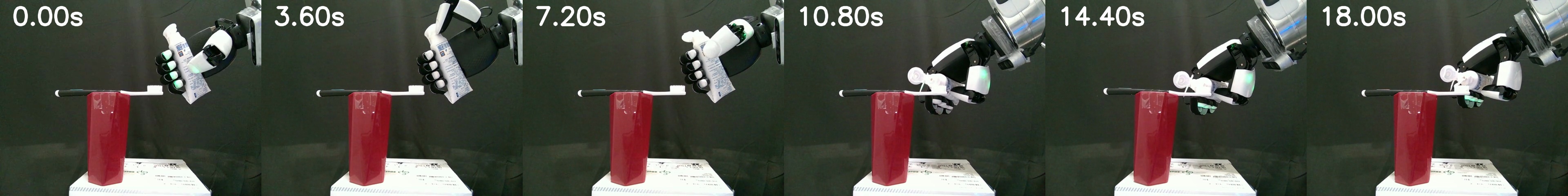}
        \caption{Toothpaste Squeezing}
    \end{subfigure}

    \begin{subfigure}[t]{1\textwidth}
        \centering
        \includegraphics[width=\linewidth]{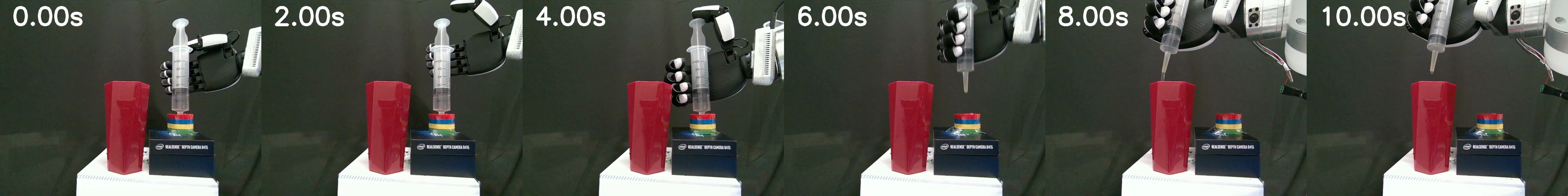}
        \caption{Syringe Pressing}
    \end{subfigure}

    \caption{Executions of trained policies on nine contact-rich manipulation tasks.}
    \label{fig:tasks}
\end{figure}

We design a suite of nine contact-rich dexterous manipulation tasks to comprehensively evaluate policy performance across diverse interaction modes. In all tasks, object poses are randomized to introduce variability in spatial configurations and contact conditions. The tasks are as follows:
\begin{itemize}[leftmargin=0.5cm]
    \item \textit{Bottle Picking}: Pick up a partially filled plastic water bottle (approximately 200g), transport it to a designated target location, and place it without dropping. This task requires stable control over a compliant object with shifting internal mass.
    \item \textit{Cup Picking}: Pick up a disposable paper cup, transport it stably, and place it at a specified location. The task emphasizes gentle contact and manipulation of deformable, lightweight objects.
    \item \textit{Egg Picking}: Pick up a raw egg, move it to a target location, and place it safely. This task demands extremely fine force modulation to prevent cracking or dropping during manipulation.
    \item \textit{Cap Twisting}: Grasp a plastic bottle, unscrew the cap, and lift it off. This task involves precise torque generation, in-hand stabilization, and coordinated finger-thumb rotation.
    \item \textit{Nut Tightening}: Rotate a plastic nut clockwise to securely fasten it onto a bolt. The task requires simultaneous rotational motion and downward force application.
    \item \textit{Peg Insertion}: Insert four wooden pegs into corresponding holes on a board. This evaluates spatial alignment, contact control, and force-guided insertion under tight tolerances.
    \item \textit{Charger Plugging}: Plug a two-prong charger into a power strip, requiring fine spatial alignment and precise force control to achieve successful insertion.
    \item \textit{Toothpaste Squeezing}: Flip open the toothpaste cap using the thumb, then squeeze paste onto a toothbrush. The task combines sequential action planning and fine-grained variable force modulation.
    \item \textit{Syringe Pressing}: Hold a syringe and press the plunger with the thumb to expel water. This simulates a one-handed, force-controlled manipulation requiring steady actuation and grasp stability.
\end{itemize}
To provide a more intuitive understanding of the tasks, we visualize the execution of the trained policies for all nine contact-rich manipulation tasks in Figure~\ref{fig:tasks}. Note that the execution times annotated in the figure may differ from those shown in Figure~\ref{fig:compare_with_teleop}, as the policies are trained with different sets of demonstrations.

\section{Policy Training Details}
\label{app:policy}

Our policy implementation builds upon the official Diffusion Policy codebase\footnote{\url{https://github.com/real-stanford/diffusion_policy}}. We retain the original architecture of the U-Net-based diffusion model and the multi-view visual encoder without modifications. However, to better support our contact-rich manipulation setting, we introduce the following enhancements:
\begin{itemize}[leftmargin=0.5cm]
    \item We integrate a tactile encoder to process high-resolution tactile inputs from five fingers. The input is a tensor of shape 5x120x3, where each of the five fingers has 120 tactile points, and each point encodes a 3D vector representing the magnitude and direction of contact force. Each finger's data is processed through a shared 1D convolutional encoder to extract per-finger features. These features are then concatenated and passed through a two-layer multilayer perceptron (MLP) to produce a fixed-length tactile embedding. This embedding is fused with visual and proprioceptive observations to form the policy input.
    \item Our policy outputs a 23-dimensional action vector, comprising 6 degrees of freedom for the end-effector pose, 12 joint angles for the dexterous hand, and 5 normal force targets at the fingertips. To improve control smoothness and temporal consistency, we adopt an action chunking strategy~\citep{zhaoLearning2023} with a chunk length of 16. During execution, we employ an interpolation controller that applies control commands at 100 Hz for the robotic arm and 50 Hz for the dexterous hand, ensuring high-frequency and stable control for both subsystems.
    \item For relatively simple tasks such as picking, plugging, and insertion, we collect approximately 100 demonstrations per task. For more challenging tasks such as \textit{toothpaste squeezing} and \textit{syringe pressing}, we collect around 150 demonstrations. All baselines are trained for 500 epochs on each task.
\end{itemize}

Some key hyperparameters for training and inference are summarized in Table~\ref{tab:training_config}.

\begin{table}[h]
\centering
\vspace{-0.5cm}
\caption{Training and inference configuration.}
\begin{tabular}{@{}ll@{}}
\toprule
\textbf{Config} & \textbf{Value} \\
\midrule
Observation horizon & 2 \\
Action horizon & 16 \\
Observation resolution & 240$\times$320 \\
Optimizer & AdamW \\
Optimizer momentum & $\beta_1, \beta_2 = 0.95, 0.999$ \\
Learning rate & 1e-4 \\
Batch size & 64 \\
Inference denoising iterations & 16 \\
Temporal ensemble steps & 8 \\
Temporal ensemble adaptation rate & -0.01 \\
\bottomrule
\end{tabular}
\vspace{-0.5cm}
\label{tab:training_config}
\end{table}

\section{Experiment Setup Details}
\label{app:exp}

\subsection{Teleoperation System Setup}
\label{app:teleop}

We replicate the Open-TeleVision~\citep{chengOpenTeleVision2024} setup and construct a single-handed teleoperation system for comparative evaluation. Our setup consists of the following components:
\begin{itemize}[leftmargin=0.5cm]
\item \textbf{Robot Platform}: A 7-DoF Franka Emika Panda robotic arm mounted with a 6-DoF Inspire Hand.
\item \textbf{Operator Interface}: A Meta Quest 3 headset is used to track the operator’s head motions in real time using its built-in hand tracking system.
\item \textbf{Motion Retargeting}: The operator’s wrist pose is mapped to the robot arm’s end-effector via a closed-loop inverse kinematics (CLIK) controller implemented with the Pinocchio library~\citep{carpentier2019pinocchio,carpentierpinocchio}, ensuring stable and precise arm control. Simultaneously, the operator’s hand keypoints are retargeted to the 12-DoF Inspire Hand using the dex-retargeting framework~\citep{qinAnyTeleop2024}, which optimizes the alignment between human and robot keypoint vectors while maintaining temporal consistency. This unified retargeting approach enables intuitive and responsive control of both the arm and the dexterous hand.
\end{itemize}

\begin{figure}[h]
    \centering
    \includegraphics[width=0.5\textwidth]{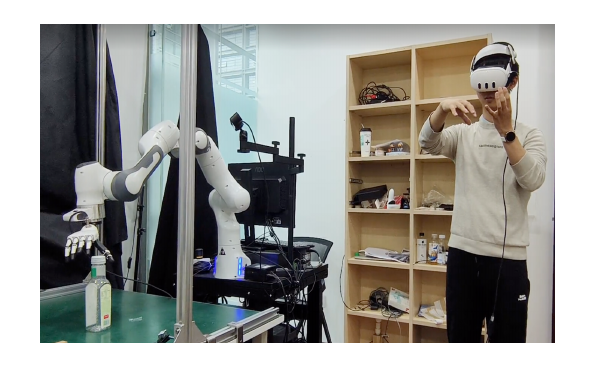}
    \caption{The overview of the teleoperation system setup.}
    \label{fig:teleop}  
\end{figure}

\subsection{User Study Setup}
\label{app:user}

We invited five participants with prior experience in robotics projects to take part in the user study. The participants had varying levels of teleoperation expertise. Each participant was guided to use both KineDex and our custom-built teleoperation system to collect demonstrations. They completed five trials on two tasks of different difficulty levels: \textit{Bottle Picking} and \textit{Syringe Pressing}. Afterward, they evaluated both systems in terms of perceived effectiveness and ease of use.

\section{Data Preprocessing Details}
\label{app:preprocess}

\begin{figure}[!t]
    \centering
    
    \begin{subfigure}[t]{1\textwidth}
        \centering
        \includegraphics[width=\linewidth]{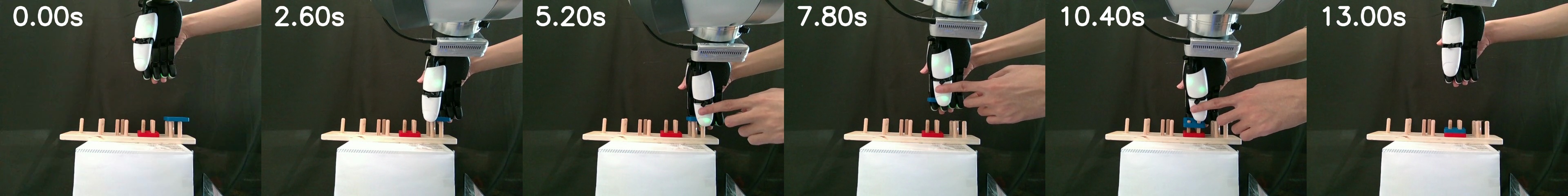}
        \caption{Kinesthetic Demonstration}
    \end{subfigure}

    \begin{subfigure}[t]{1\textwidth}
        \centering
        \includegraphics[width=\linewidth]{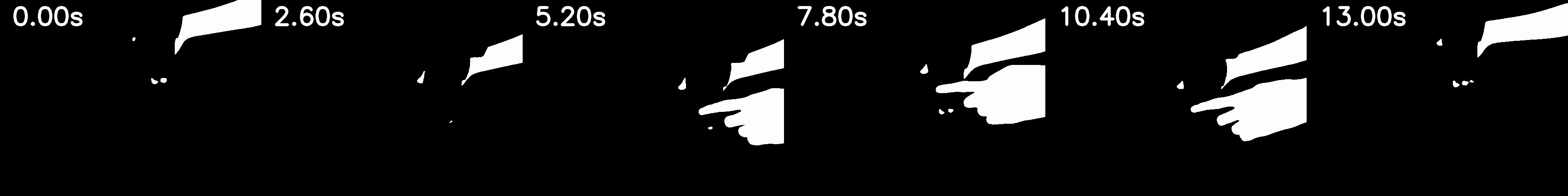}
        \caption{Masked Demonstration.}
    \end{subfigure}

    \begin{subfigure}[t]{1\textwidth}
        \centering
        \includegraphics[width=\linewidth]{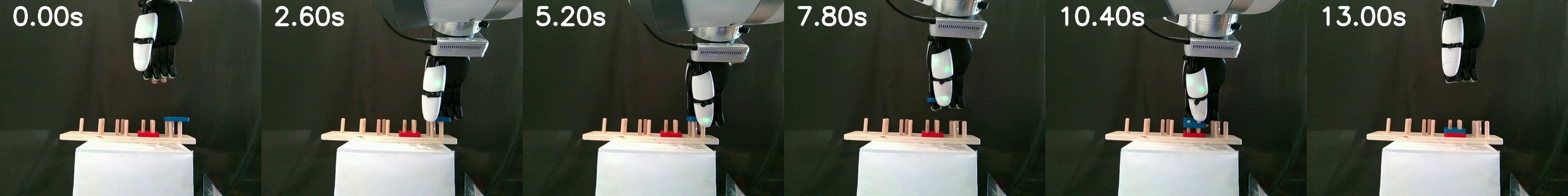}
        \caption{Inpainted Demonstration.}
    \end{subfigure}

    \caption{Data preprocessing pipeline for \textit{Peg Insertion}.}
    \label{fig:preprocess1}
\end{figure}

\begin{figure}[!t]
    \centering
    
    \begin{subfigure}[t]{1\textwidth}
        \centering
        \includegraphics[width=\linewidth]{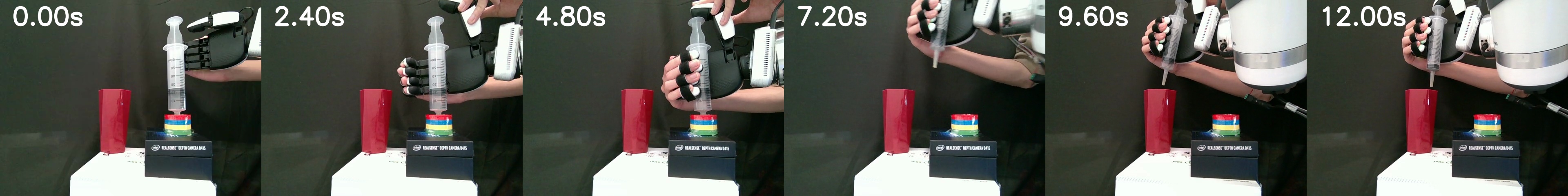}
        \caption{Kinesthetic Demonstration}
    \end{subfigure}

    \begin{subfigure}[t]{1\textwidth}
        \centering
        \includegraphics[width=\linewidth]{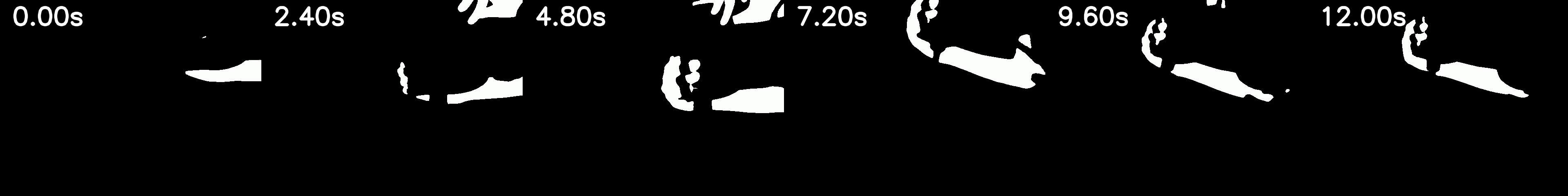}
        \caption{Masked Demonstration.}
    \end{subfigure}

    \begin{subfigure}[t]{1\textwidth}
        \centering
        \includegraphics[width=\linewidth]{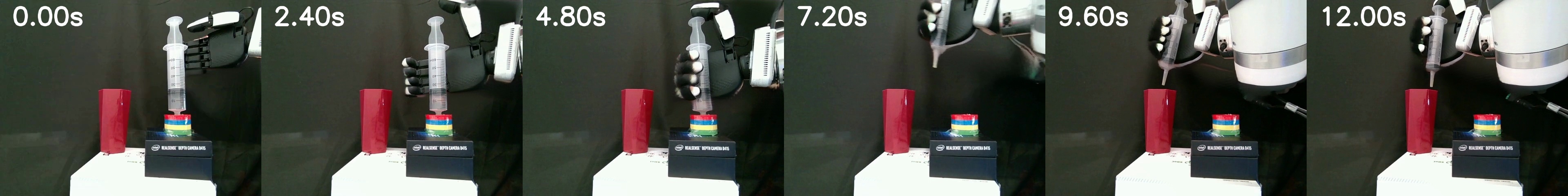}
        \caption{Inpainted Demonstration.}
    \end{subfigure}

    \caption{Data preprocessing pipeline for \textit{Syringe Pressing}.}
    \label{fig:preprocess2}
\end{figure}

The KineDex data preprocessing pipeline consists of three stages. First, demonstrations are collected via kinesthetic teaching. Second, we use Grounded-SAM~\citep{renGrounded2024} to segment human body regions and generate corresponding masks. Finally, we use ProPainter~\citep{zhouProPainter2023} to inpaint the occluded areas by removing the masked regions. Two examples are illustrated in Figures~\ref{fig:preprocess1} and~\ref{fig:preprocess2}.

\end{document}